%% file: LNAS_main.tex
\documentclass{article} 
\usepackage{iclr2021_conference,times}

\input{math_commands.tex}

\usepackage{hyperref}
\usepackage{url}
\usepackage{graphicx}
\usepackage{subfigure}
\usepackage{graphics}
\usepackage{caption}
\title{Generalized Latency Performance \\Estimation for
Once-For-All \\Neural Architecture Search}

\author{Muhtadyuzzaman Syed \& Arvind Akpuram Srinivasan\\
Georgia Institute of Technology\\
Atlanta, GA 30332, USA \\
\texttt{\{msyed32,asrinivasan87\}@gatech.edu} \\}

\iclrfinalcopy 
\begin{document}

\maketitle

\begin{abstract}

Neural Architecture Search (NAS) has enabled the possibility of automated machine learning by streamlining the manual development of deep neural network architectures defining a search space, search strategy, and performance estimation strategy. To solve the need for multi-platform deployment of Convolutional Neural Network (CNN) models, Once-For-All (OFA) proposed to decouple Training and Search to deliver a one-shot model of sub-networks that are constrained to various accuracy-latency trade-offs. We find that the performance estimation strategy for OFA's search severely lacks generalizability of different hardware deployment platforms due to single-hardware latency lookup tables that require significant amount of time and manual effort to build beforehand. In this work, we demonstrate the framework for building latency predictors for neural network architectures to address the need for heterogeneous hardware support and reduce the overhead of lookup tables altogether. We introduce two generalizability strategies which include fine-tuning using a base model trained on a specific hardware and NAS search space, and GPU-generalization which trains a model on GPU hardware parameters such as Number of Cores, RAM Size, and Memory Bandwidth. With this, we provide a family of latency prediction models that achieve over 50\% lower RMSE loss as compared to with ProxylessNAS. We also show that the use of these latency predictors match the NAS performance of the lookup table baseline approach if not exceeding it in certain cases. 



\end{abstract}

\section{Introduction}

Model deployment of Convolutional Neural Networks (CNN) is an emerging topic, particularly for edge devices. As applications for image processing and object recognition tasks continue to grow, the need for tailored models with the highest possible level of accuracy is paramount for near human-like performance. Diverse sets of hardware platforms have different resource constraints defining unique latency and workload requirements. Traditionally for each deployment scenario, Neural Network Architectures are manually designed and trained to meet deployment constraints such as latency. This becomes computationally prohibitive with respect to design time, expertise, GPU hours, and $CO_2$ emission as the number of deployment scenarios scale exponentially. 

Once-For-All (OFA) (Cai et al., 2020 [1]) proposed to streamline the model deployment process by introducing the Once-For-All network that contains sub-networks with unique accuracy-latency trade-offs to serve $10^{19}$ deployment scenarios at once. OFA achieves this by decoupling Training and Neural Architecture Search (NAS), which reduced training cost from a per scenario basis to a one time cost of 1200 GPU hours. 

Focusing on OFA's search paradigm, NAS can be categorized into three dimensions: search space, search strategy, and performance estimation strategy (Elsken et al., 2019 [2]). OFA defines the search space of CNN child architectures with varying depth, width, kernel size, and image resolution presenting over $10^{19}$ possible model configurations for deployment. Compound Once-For-All (CompOFA) (Work-In-Progress) establishes a several orders of magnitude smaller search space of 243 model configurations and eliminates the need for training sub-optimal models below the accuracy-latency Pareto frontier, establishing half the cost of GPU hours, dollars, and $CO_2$ emission for training. 

Performance Estimation is the process of estimating predictive performance of architectures found by NAS during search. For example, with a search strategy defined as an evolutionary search algorithm, performance estimation serves as a feedback mechanism for NAS to indicate whether the child model found is optimal {\it without} the need for retraining. OFA defines their performance estimation strategy as individual latency lookup tables specified for a single hardware deployment platform. Each lookup table contains a manual enumeration of every possible layer in a CNN architecture and its corresponding latency estimation on that hardware platform. The latency estimation for each layer is calculated by measuring over some number of iterations and simply taking the mean. Further details of how the lookup tables are created are not shared by the original authors and leads to inefficiency in latency guided NAS for heterogeneous hardware as manual calculation of latency are required for hardware which do not have look-up tables. 

In this work, we focus primarily on developing a generalized performance estimation strategy to efficiently perform latency guided NAS. We have developed our work on top of OFA and CompOFA in order to serve a working proof of concept that machine learning latency prediction models {\it can} be used generally for latency-guided NAS applications.\\ \\
We achieve the following goals in this work: \\
(1) Develop latency predictors for CNN architectures with SOTA performance\\
(2) Enable heterogeneous hardware support through 2 generalization strategies: fine-tuning and GPU-generalization\\
(3) Reduce the manual overhead of creating lookup tables in the baseline approach proposed by OFA\\


\section{Related Work}
\label{gen_inst}

\textbf{Model-based performance estimation for NAS:} To reduce the computation and time taken to evaluate the performance of a model during search, performance estimators have been employed. In Cai et al. [1], an accuracy predictor is used for predicting the accuracy of a given model. A lookup table is used for calculating the latency of a model and contains the latencies for every possible layer in a CNN architecture. The total latency is found by simply taking the sum total of all individual layers. In Cai and Han. [3], the latency of a model is predicted using a latency predictor that takes into account layer operation, depth, and width among other parameters to estimate the latency.

\textbf{GPU performance estimation:} GPU performance estimation focuses on calculating the performance and/or energy consumption of a GPU hardware based its configuration. (Note that this is not in the context of NAS for a performance estimation strategy.) GPU performance can be estimated from equations that have memory frequency and number of cores as variables (Figuera and Issa[6], Wang and Chu[4]). Paul et al.[5] has shown that performance estimation can be dependent on the memory size and memory hierarchy that is present in a GPU.

\section{Motivation}
\label{headings}

\subsection{Latency Estimation for NAS}

For latency performance estimation, there are two methods used in the NAS literature. One is to follow a lookup table approach used by OFA and the second is to train a latency predictor used by ProxylessNAS. Unfortunately, ProxylessNAS has not released their code base or any information about their latency predictors, therefore we can only rely on their statements on how they perform. We additionally do not know how they are implemented or how the performance fares across different types of hardware. In this work, we create a family of latency predictors with a focus on hardware generalization and examine their performance in a variety of deployment scenarios.

\subsection{Generalized Latency Estimation across Hardware}

In the case of both OFA and ProxylessNAS, the estimation strategies must be applied for each hardware individually. The main advantage of using a latency prediction model is that it could be tailored for heterogeneous hardware support by simply making modifications on the data it is trained on. This ultimately helps to reduce the manual overhead of creating lookup tables for each hardware scenario. We aim to offer the framework for developing hardware generalized prediction models for the purposes of accelerating research in latency-guided NAS applications.

\subsubsection{Capturing the Effect of Model Architecture on Latency}

The latency of a model is based on the model architecture and the hardware parameters. Since the model architecture will remain the same across different hardware, the underlying effect of the model architecture will allow for some level of generalization of a model across hardware.

Additionally, given a set of architectures, their ordering in terms of latency may in certain cases remain the same across different types of hardware. This could be the case when looking at different architectures in the OFA search space and therefore allows for the underlying ordinality to be captured for generalization (eg: Irrespective of the hardware in use, increasing the depth of a model will increase the latency of the model). 

It would be remiss to not mention that this will not always be the case. With the advent of highly specialized hardware that is created with particular configurations in mind, this assumption will break (eg: CNN specialized hardware may have lower latency for CNN models when compared to similarly large LSTM models, which need not be the case if a general GPU is used). This is not likely to be the case in OFA, because the models while differing in terms of depth, width and kernel size are all similar in terms of layer operations (eg: convolution layer).

\subsubsection{Parametrization of Hardware}

The effect of a hardware on the latency is the result of the configuration of the hardware in use. If the hardware can be parametrized in terms of its configuration, we can capture the underlying effect of a configuration on the latency of a model, thereby not assuming it simply as a black box and therefore not necessitating an individual estimator for each new hardware in question. 

\begin{figure}[t]
\centering
\includegraphics[width=13.9cm]{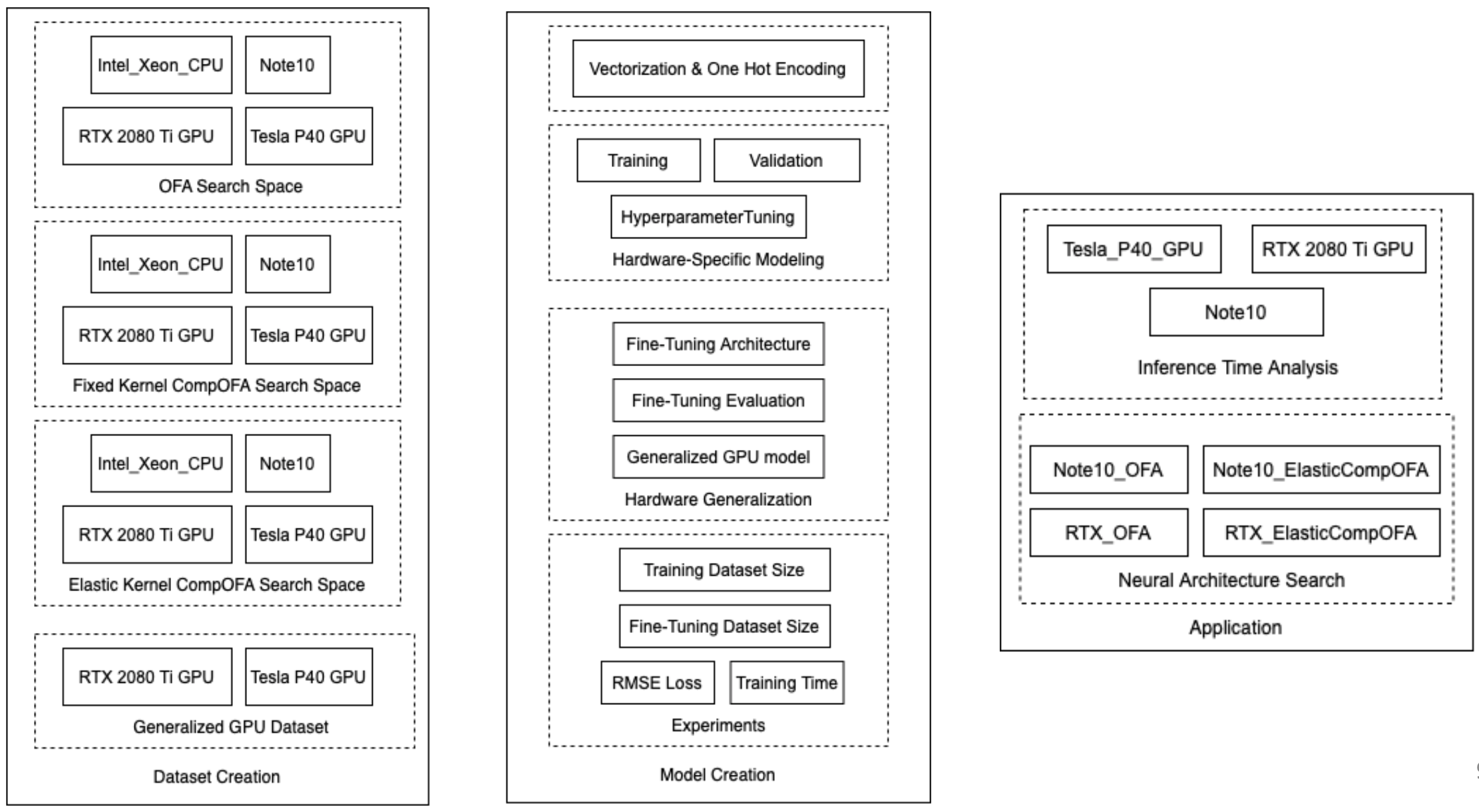}
\caption{High level design of latency prediction model development consisting of Dataset Creation, Model Creation, and Application. }
\label{fig:design}
\end{figure}

\section{Latency Performance Estimation}

The high level design of this work can be visualized in Figure \ref{fig:design}, which consists of three stages: Dataset Creation, Model Creation, and Application. In Dataset Creation, we generate 10 datasets of (child\_architecture, latency) pairs in various combinations of hardware deployment platforms and NAS search spaces. In Model Creation, we develop 2 groups of predictors: hardware-specific models and hardware-generalized models. We use techniques such as vectorization, One Hot Encoding, Hyperparameter Tuning, Fine-Tuning, and Model Evaluation. In Application, we assess the usage of the latency predictors in an existing NAS application to see how it performs compared to baseline approaches. 

\subsection{Dataset Creation}
To develop the generalized latency predictors, we generate 8 datasets of (child\_architecture, latency) pairs for the purpose of model training and testing. We also generate 2 additional datasets for GPU generalization.

\subsubsection{Architecture Search Spaces}
As designed by OFA, we consider search spaces covering 4 dimensions of CNN architectures which include kernel size, depth, width, and image resolution. With these parameters, each search space defines the total number of different child architectures that are possible. For example, for an OFA search space defining parameter elasticity with 3 layers, 3 channels, 3 kernel sizes, and an image size with a unit size of 5, the total number of model configurations is (3 x 3)$^2$ + (3 x 3)$^3$ + (3 x 3)$^4$)$^5$ = $10^{19}$ models. Another search space we consider in our latency predictor evaluation is CompOFA. The CompOFA search space aims to shrink the design space of the model architectures in order to avoid considering sub-optimal models along the accuracy-latency curve. This reduces GPU hours and CO2 emission during the Training stage, and proves to generate models that perform on-par with models from OFA.  By incorporating a Coupling Heuristic, CompOFA defines two possible search spaces: Fixed kernel and Elastic kernel. Fixed kernel creates a simplified space with a kernel size of either 3 or 5 in each block, generating 3$^5$ = 243 models. For Elastic kernel, kernel sizes of \{3,5,7\} are considered for every layer, generating (3$^2$ + 3$^3$ + 3$^4$) = 10$^{10}$ models. We refer to these search spaces as "OFA", "Fixed-Kernel CompOFA", and "Elastic-Kernel CompOFA".

\subsubsection{Latency on Hardware Platforms}
We consider multiple hardware platforms for the aim of hardware generalization. These platforms include the Samsung Galaxy Note 10, Intel Xeon CPU, RTX 2080 Ti GPU, and Tesla P40 GPU. Considering a child architecture from a search space as defined previously, each model will contain a different latency measurement on each hardware. For example, we can consider a sample architecture from Fixed-Kernel CompOFA as the following: 
\begin{center}
 \{'ks': [3, 3, 3, 3, 5, 5, 5, 5, 3, 3, 3, 3, 3, 3, 3, 3, 5, 5, 5, 5], \\
  'e': [6, 6, 6, 6, 4, 4, 4, 4, 4, 4, 4, 4, 6, 6, 6, 6, 6, 6, 6, 6], \\
  'd': [4, 3, 3, 4, 4], \\
  'r': [176]\}
\end{center}

When performing latency measurement on the RTX 2080 Ti GPU, this model architecture has a latency of 18.66242886 milliseconds (ms). For the same model measured on the Intel Xeon CPU, the latency is 1001.928711 ms. This variance makes sense due to the differences in computation power between a CPU and GPU during model training and inference. For each hardware platform, we create datasets for each search space. Two more datasets were created for training a latency predictor that is generalized over GPUs. To account for GPU characteristics, additional heuristics were added such as Number of Cores, RAM size, and Memory Bandwidth. More details on why these specific parameters were chosen is explained in Section 4.5.

The datasets used for training the latency predictors were created by randomly sampling child architectures found in each respective search space [OFA, Fixed-Kernel-CompOFA, Elastic-Kernel-CompOFA]. For each child architecture sampled, latency measurement is performed and recorded. Each datapoint is assumed to be independently and identically distributed (iid) offering a wide enough variety of model possibilities that can be used for training. Each dataset was generated with 10,000 datapoints of (child\_architecture, latency) pairs where 70\% is used training, 15\% for testing, and 15\% for validation during model creation.

A limitation to note is that the Fixed-CompOFA search space does limit the number of model configurations possible to only 243. For the purposes of training the latency predictors, we want to produce as many unique model configurations as possible so our predictors can learn generally over various architectures. From our experiments, limiting to the Fixed-Kernel-CompOFA search space led to overfitting which does not aid in accurate latency prediction in the wild. Hence, we use models trained in OFA to honestly evaluate performance in our experiments.

\begin{figure}[t]
\centering
\includegraphics[width=5cm]{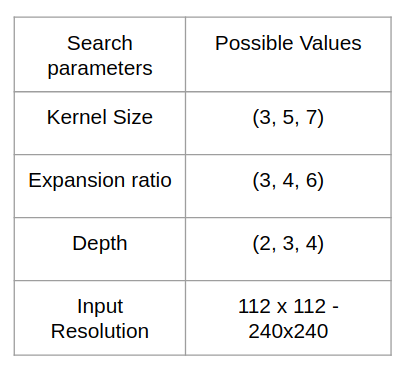}
\caption{Search space of OFA}
\label{fig:search_space}
\end{figure}

\subsection{Model Creation}
\subsubsection{Vectorization}

The model configurations in the OFA search space need to be parametrized and vectorized so that it can be used as an input for the latency predictors. We have chosen to follow the vectorization strategy that OFA used for their accuracy predictor. The accuracy predictor has a Root Mean Squared Error (RMSE) loss of 0.21\%, and is hence able to predict the accuracy simply given the architecture. Since the latency estimation is also based on the architecture of the model, the assumption is that this vectorization process will work for a latency predictor as well. In the case of a generalized latency predictor, where the hardware configuration is also considered, the input vector will have some modifications, as explained in section 4.3.

The OFA search space is characterized by the kernel size of a layer (3 possible values), the expansion ratio of a layer (3 possible values), the depth of a block (3 possible values), and the input image resolution (refer Figure \ref{fig:search_space}). There are 5 blocks in every model and the maximum possible depth value is 4, therefore the maximum number of layers is 20. The vectorization strategy is as follows:

\begin{enumerate}
    \item The kernel size (\textit{ks}) and expansion ratio (\textit{e}) of each layer is one-hot encoded. (If the layer does not exist, because the depth is less than 4, then its corresponding encoding is (0,0,0))
    \item The encoding for \textit{ks} and \textit{e} are concatenated to form two vectors of length 60 each (20 (maximum possible layers) x 3 (length of each encoding))
    \item The possible input image resolutions (\textit{r}) are between $112 \times 112$  and $240 \times240$. The \textit{r} is encoded into a one-hot vector of length 8. The interval between 112 and 240 is divided into 8 equally spaced intervals and each bit represents one of the intervals.
    \item The final vector is a concatenation of the vectors representing \textit{ks, e, r}. Therefore the input vector is of length 60 + 60 + 8 = 128. 
\end{enumerate}

\subsubsection{Model Training}

\begin{figure}
\hfill
\subfigure[Model configuration]{\includegraphics[width=5cm]{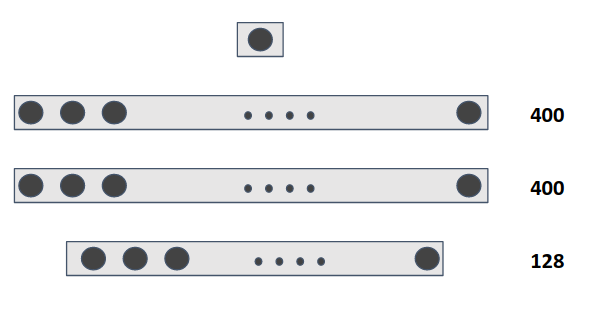}}
\hfill
\subfigure[Model parameter - search space]{\includegraphics[width=5cm]{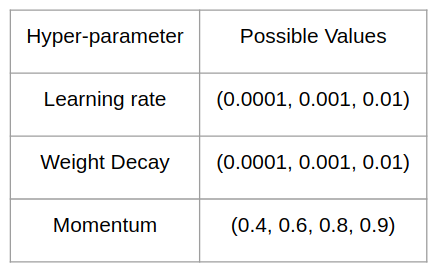}}
\hfill
\caption{Latency Predictor}
\label{fig:Latency_predictor}
\end{figure}

We use a deep learning model architecture (Figure \ref{fig:Latency_predictor} (a)) for our latency predictors. It is a 3 layer feed-forward network with 400 neurons in the inner 2 layers and 1 neuron in the output representing the latency prediction value. The architecture is similar to that used for the accuracy predictor by OFA. The loss function is Root Mean Squared Error (RMSE) and each model is trained for 100 epochs with a batch size of 32. The data split is 70\% training, 15\% testing, and 15\% validation. There are three hyper-parameters used for training, namely: Learning Rate, Weight Decay and Momentum (Figure \ref{fig:Latency_predictor} (b)). During training, we perform hyper-parameter tuning using RayTune (Liaw et al., 2018 [7]), which provides a distributed computation framework for machine learning model development. We implement a grid search for each model hyper-parameter using Asynchronous Successive Halving (ASHA) as our trial scheduler, ultimately training 36 different models at once in a reasonable time frame. We train two groups of latency predictors. Hardware-specific models that are trained directly on data from a single hardware, and Hardware-generalized models that are fine-tuned to support multiple hardware platforms with little overhead to re-train as compared to training from scratch.

\subsubsection{Fine-Tuning}

To achieve hardware generalization, where a latency predictor trained on a single hardware can be made to predict for other kinds of hardware, we utilize fine-tuning (Han et al., 2015 [8]). Our approach to fine-tuning is to assume that the lower layers of a latency predictor architecture are more likely to learn general features that may apply to all types of hardware. This assumption may not hold in cases of hardware where the latency distribution is significantly different from the original hardware, such that the general features learnt from the initial layers do not hold true anymore.

To fine-tune, we freeze the first two layers of the model and train on a data set size of 700, a data set size which is 80\% smaller when compared to training from scratch. Justification for the fine-tune data set size is provided in the experiments section 5.2 .

\subsection{GPU Latency Generalization}

{\it Complete} generalization on the basis of hardware can be understood as the ability of a model to predict the latency of an architecture for a {\it new} hardware, without additional training. To achieve this type of generalization, we would need to parametrize the hardware itself based on its configuration so that a reasonable latency prediction can be made. Generalization across hardware such as CPUs, GPUs and mobile hardware units is very difficult due to the fundamental differences they have in their respective configurations. 

Therefore, we focused only on the family of GPU hardware for generalization to show a proof of concept that can be potentially extended for other hardware platforms. To choose the possible parameters that can be used to represent a GPU, we reviewed literature on GPU performance estimation ({\it Note: not in the context of NAS}), and chose three widely cited parameters - Number of Cores, RAM size and Memory Bandwidth. The input vector of the latency predictors is changed accordingly by appending the numerical values of these three parameters, thereby increasing the length from 128 to 131. The assumption here is that parameters that affect GPU performance in computer programs and games, can also explain the hardware related latency effects for a model in the NAS paradigm.

\section{Experiments}

\begin{table}
\centering
\begin{tabular}{||c| c| c ||}
\hline
Model&RMSE Loss abs.(ms)&RMSE Loss \% (ms \%)\\
\hline\hline
Proxyless NAS & 0.75 & -\\
Latency Predictor - Note 10 & 0.325 & 1.02\%\\
Latency Predictor - RTX GPU & 0.385 & 2.16\%\\
Latency Predictor - Tesla GPU & 1.18 & 2.5\%\\
Latency Predictor - Intel Xeon CPU & 14 & 1.7\%\\
\hline
\end{tabular}
\caption{Test Loss for Latency Predictors}
\label{table:ind_table}
\end{table}

\begin{figure}
\hfill
\subfigure[Loss Curve]{\includegraphics[width=6cm]{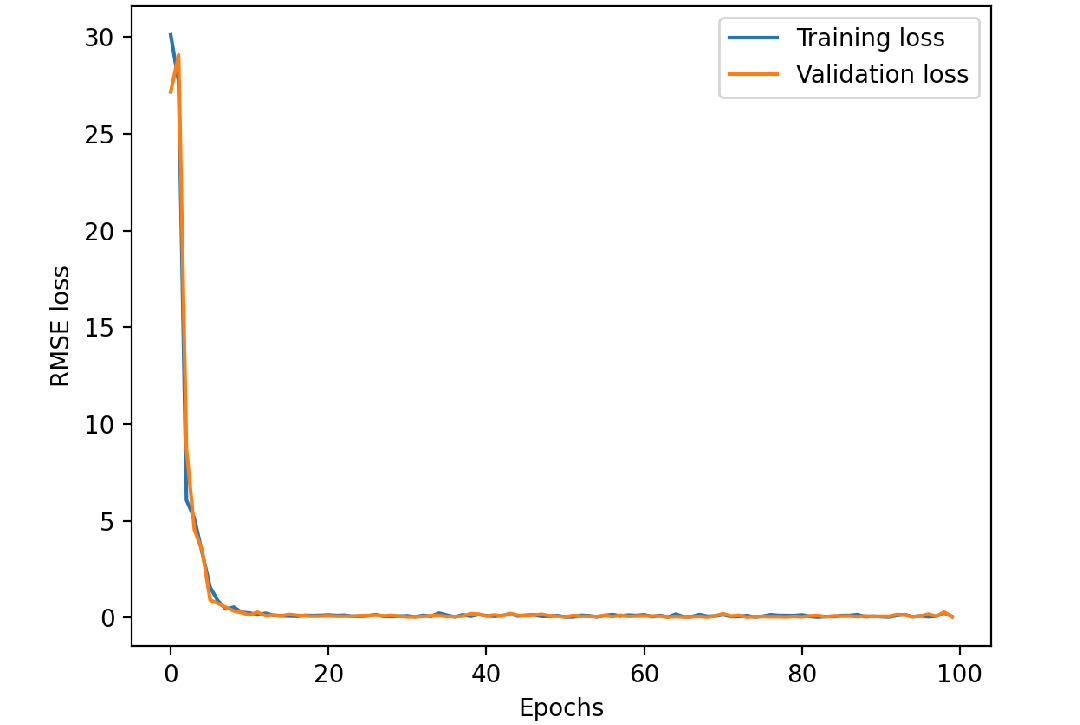}}
\hfill
\subfigure[Effect of training data set size]{\includegraphics[width=6cm]{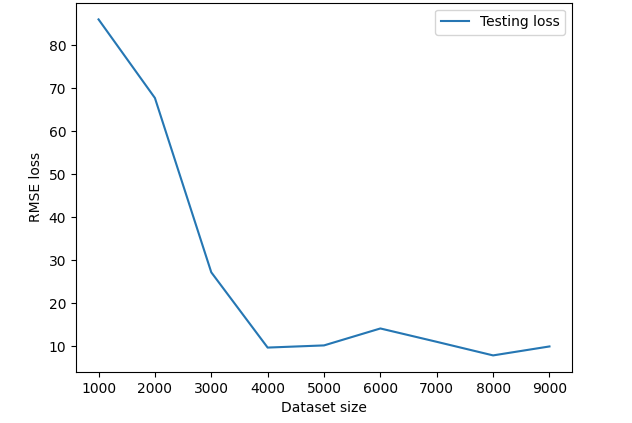}}
\hfill
\caption{Latency Predictor model training}
\label{fig:Ind_model_training}
\end{figure}

\subsection{Test Loss of Predictors}

We trained latency predictors for 4 different hardware - Samsung Galaxy Note 10, Intel Xeon CPU, RTX 2080 Ti GPU, and Tesla P40 GPU. In addition to RMSE, we also considered RMSE \% which is RMSE/(avg. latency of models for a hardware), since the range of latencies across hardware varied significantly. For example, CPUs have a range of 500-1000 ms whereas GPUs are usually in the range of 20-50 ms. This metric would enable better comparison between different hardware. As seen in Table \ref{table:ind_table}, the RMSE loss of our predictor performs either equal or superior to that of ProxylessNAS. Figure \ref{fig:Ind_model_training} (a) shows an example of the convergence of the loss curve of a latency predictor and Figure \ref{fig:Ind_model_training} (b) shows that the optimal training dataset size is 4000 when identifying the "elbow" in the loss curve.

\begin{figure}
\hfill
\subfigure[Loss Curve]{\includegraphics[width=6cm]{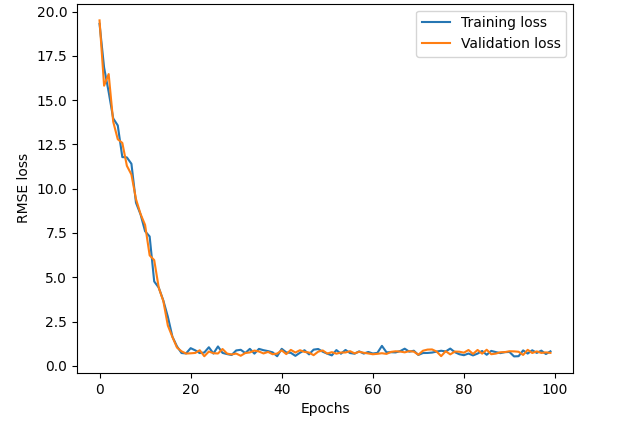}}
\hfill
\subfigure[Effect of fine-tune data set size]{\includegraphics[width=6cm]{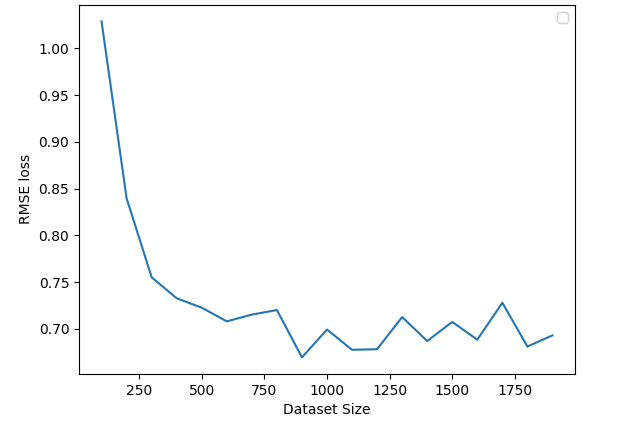}}
\hfill
\caption{Latency Predictor model fine-tuning}
\label{fig:Fine-tune_model_training}
\end{figure}

\begin{table}
\centering
\begin{tabular}{||c| c | c ||}
\hline
HW to be fine-tuned for & Original Model RMSE (ms)&Fine-tuned Model RMSE(ms)\\
\hline\hline
RTX GPU & 0.385 & 0.712\\
Tesla P40 GPU & 1.181 & 1.609\\
Intel Xeon CPU & 14 & 171\\
\hline
\end{tabular}
\caption{Test Loss for fine-tuned latency predictors (using Note10 base model)}
\label{table:finetune_table_Note10}
\end{table}

\begin{table}
\centering
\begin{tabular}{||c| c | c ||}
\hline
HW to be fine-tuned for & Original Model RMSE (ms)&Fine-tuned Model RMSE(ms)\\
\hline\hline
Note 10 & 0.3525 & 7.45\\
Tesla P40 GPU & 1.181 & 10.82\\
Intel Xeon CPU & 14 & 313\\
\hline
\end{tabular}
\caption{Test Loss for fine-tuned Latency Predictors (using RTX GPU base model)}
\label{table:finetune_table_RTX}
\end{table}

\subsection{Test Loss of Fine tuning}

We took a fully trained Note10 latency predictor and fine-tuned it to see if it would generalize to other hardware. As seen in Table \ref{table:finetune_table_Note10}, the RMSE from the hardware-specific models and a fine-tuned models are comparable, specifically for the RTX and Tesla P40 GPUs. 

It however does not work for the Intel Xeon CPU, where the RMSE loss of the fine-tuned model is substantially larger. This is likely due to the range of latencies of the CPU (500-1000ms) which is very different from the range of latencies of the Note10 (20-50ms), and therefore, freezing the first two layers hampers the predictive performance. 

Figure \ref{fig:Fine-tune_model_training} (a) shows an example of the loss curve convergence for the fine-tuned latency predictor of the RTX GPU (using the Note10 base model) and Figure \ref{fig:Fine-tune_model_training} (b) shows that the optimal fine-tune dataset size is 700.

We wanted to see if performance for fine-tuning was dependent on the base model chosen. So we also tried fine-tuning using the RTX GPU base model. As can be seen from Table \ref{table:finetune_table_RTX}, the RMSE loss of the fine-tuned models is significantly higher than that of the original models. Therefore, fine-tuning on the Note10 base model gave superior results as compared to using the RTX GPU base model. This implies that the base model chosen {\it does} affect the performance of fine-tuning.

\begin{figure}
\centering
\begin{minipage}{.5\textwidth}
  \centering
  \includegraphics[width=7cm]{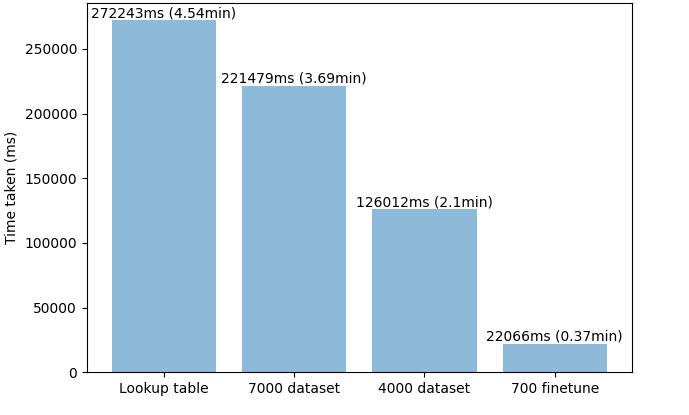}
  \caption{Time Overhead}
  \label{fig:time_taken}
\end{minipage}%
\begin{minipage}{.5\textwidth}
  \centering
  \includegraphics[width=7cm]{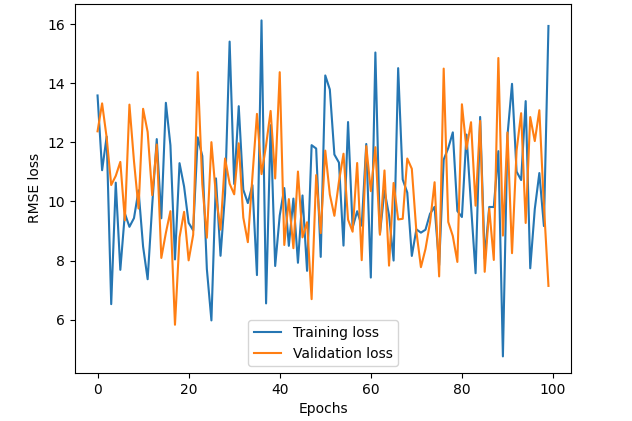}
  \caption{GPU generalized Loss curve }
  \label{fig:gen_gpu}
\end{minipage}
\end{figure}

\subsection{Time Taken to Create Lookup Table vs Dataset}

The main goal of this work has been to see if generalization of latency prediction across hardware is possible. We thought it would be prudent to also show a comparison of time taken to create a dataset for training as compared to the overhead of creating a lookup table. This would provide insight of how easy it is to create a latency predictor compared to a lookup table.

To calculate the time taken to create a training dataset, we simply calculated the total latencies of all the models in the training dataset (as we have to manually run a model to calculate the latency of that model). 

The lookup table is an enumeration of all the possible layers that a family of models can have along with the latency of each of the layers. The time to create a lookup table was hence simply the sum of all the latencies of all the possible layers. This is excluding any other time estimates which the authors of OFA have not disclosed for lookup table creation. An assumption here is that it takes significantly more time to create a lookup table, keeping in mind the time for manual enumeration and/or time to create an automated program for lookup table creation in a YAML structure, as the authors have provided.

As can be seen from Figure \ref{fig:time_taken}, the training dataset (optimal size is 4000) takes half the amount of time to create compared to a lookup table, and the fine-tune dataset (optimal size is 700) takes 12x lesser time.

\begin{table}
\centering
\begin{tabular}{||c| c ||}
\hline
Training type & Time Taken (s)\\
\hline\hline
Model Training & 78\\
Model training (Hyperparameter tuning) & 1660\\
Model Fine-tuning  & 11\\
\hline
\end{tabular}
\caption{Model training time}
\label{table:model_training_time}
\end{table}

\subsection{Model training time}

Table \ref{table:model_training_time} shows the amount of time it takes to train a latency predictor. As can be seen from the table, it is quite small. A hyperparameter tuned model take 1660 s, because we are performing grid search across 36 different types of models (as can be seen from Figure \ref{fig:Latency_predictor} (b)). Please do note that the training has been done on laptop CPUs, so these are upper range values, if performed on a GPU, the training time would be substantially lower. 
 
\subsection{Test Loss of Generalized GPU latency predictor}

To test the performance of the GPU generalized latency predictor described in section 4.3. The training data set was composed of data from two different GPUs (RTX 2080 and Tesla P40) and was trained on a data set of size 14000. As can be seen from Figure \ref{fig:gen_gpu}, the loss curve does not converge. The RMSE loss comes to about 7-11ms.

We believe that the latency predictor is going into mode collapse and simply choosing the average of the latency amongst the two hardware. This is likely occurring for two reasons, the first is that the data set contains data only from two different types of GPUs. If instead it had data from a multitude of different GPUs (of different configurations), the generalization could have been better. The second reason is that parameters for the hardware in the input vector are numerical, while the encoding for the architecture is one-hot. This dissonance could be causing some issues as well.

\subsection{Application in Neural Architecture Search}
After training the latency predictors, we apply them to an existing NAS framework to evaluate performance and the extensibility of model-based performance estimation for application in search. We analyze the differences in search time and accuracy-latency trade-offs with our approach compared with baseline approaches used in OFA and CompOFA. The goal here is not to claim that our approach performs {\it better} than other approaches but rather, we take an objective look at {\it utilizing} our latency predictors in NAS applications. We hope that our approach at the least does not cause any performance degradation in NAS. We realize that in the context of OFA, reducing search time does not provide significant benefit in terms of reducing overall cost, as the bottleneck comes from the training stage. We are simply evaluating an application of our latency estimation strategy in various deployment scenarios and seeing how our latency generalization fairs as a tradeoff with model accuracy, latency, and search time.

\begin{table}
\centering
\begin{tabular}{||l|r|r||}
\hline
\textbf{Hardware} & \textbf{Latency Estimation Strategy} & \textbf{Inference Time (ms)}\\
\hline\hline
Tesla P40 GPU & RTX\_OFA\_model & 0.564\\
              & Note10\_OFA\_model & 0.544\\
              & Latency\_Measurement & 1195.303\\
\hline
RTX 2080 Ti GPU & RTX\_OFA\_model & 0.277\\
                & Note10\_OFA\_model & 0.274\\
                & Latency\_Measurement & 706.820\\
\hline
Intel Xeon CPU & RTX\_OFA\_model & 6.236\\
               & Note10\_OFA\_model & 5.977\\
               & Latency\_Measurement & 6807.215\\
\hline
\end{tabular}
\caption{Inference time analysis of latency estimation strategies on 3 hardware platforms}
\label{table:inference_time}
\end{table}

To first get a sense of how long a single latency estimation takes, we compare inference times of two trained latency predictors with the latency measurement method used in CompOFA (See Table \ref{table:inference_time}). We evaluate on three different hardware platforms including the Tesla P40 GPU, RTX 2080 Ti GPU, and Intel Xeon CPU. We notice that the inference time for a single forward pass on both latency predictors is significantly faster as compare to latency\_measurement.

In Appendix \ref{Appendix:nas}, Figures \ref{fig:nas_application} (a) and (b), we perform NAS for Samsung Galaxy Note 10 using the Fixed-Kernel (243 models) and Elastic-Kernel ($10^{10}$ models) CompOFA search spaces. We search for models in the Once-For-All network for 5 latency constraints: 15, 20, 25, 30, 35 ms comparing 3 latency estimation strategies including the lookup table, Note10 trained latency predictor, and Note10 latency predictor with memoization. Memoization was added as an additional component to search in order to save progress on child architectures found. During the search, if we found an architecture we have seen before, we simply lookup the corresponding latency estimate rather than using the latency prediction model. This speeds up search time for smaller search spaces such as Fixed-Kernel. 

For the search, we use an evolutionary algorithm, defining a population size of 100 and number of generations as 500. As seen in the figures, our note10 model with memoization consistently, on average, finds model architectures with higher Top-1 and Top-5 ImageNet accuracy than the other two approaches and lower search time only in the case of Fixed-kernel. For Elastic-kernel, the lookup table approach performs the fastest, albeit with less accurate models. We can attribute this to the fact that memoization simply does not help as much for larger search spaces as it is less likely that a model has been seen more than once during search. In this case, there is a tradeoff between model accuracy and NAS search time. Our latency predictor just in itself, performs on-par.

In Appendix \ref{Appendix:nas}, Figures \ref{fig:nas_application} (c) and (d), we perform the same experiment with the RTX 2080 Ti GPU, comparing the latency measurement approach from CompOFA, the trained latency predictor on the RTX GPU, and the fine-tuned RTX model using the Note10 base model. We can see that on average, the fine-tuned model performs on-par with the other two approaches, exceeding them for some latency constraints. Our goal here is to show that there is no performance degradation when using our fine-tuned models. They perform similarly to baseline approaches for OFA NAS, keeping in mind the offered benefits of reducing lookup table overhead and most importantly, hardware generalization.


\section{Conclusion}

In this work, we have shown a framework for performing latency estimation using neural networks for models belonging to the Once-For-All Neural Architecture Search paradigm. We have proposed two hardware generalization strategies which include fine-tuning using a base model trained on a specific hardware, and GPU-generalization which trains a model to generalize over different GPUs by taking as input, GPU hardware parameters such as Number of Cores, RAM size, and Memory Bandwidth.

We generate and test our latency predictors over a diverse set of hardware namely - Samsung Galaxy Note 10, Intel Xeon CPU, RTX 2080 Ti GPU, and Tesla P40 GPU to test their applicability. We show that our latency predictors achieve up to 50\% lower RMSE as compared to that mentioned in ProxylessNAS.

We also show that the fine-tuned models can achieve comparable results with hardware-specific models, while being trained on datasets that are about 80\% smaller than the training dataset size. We therefore illustrate that fine-tuning, as a generalization strategy, significantly reduces the overhead of lookup table creation and hence makes it an appealing approach for hardware generalization. While our results for GPU generalization have significant room for improvement, it introduces a methodology that can be used for future research and when larger and more diverse datasets are available. 

Finally, we applied our latency predictors to assess their performance in NAS applications and found that both hardware-specific and hardware-generalized models perform on-par with other approaches such as lookup tables and latency measurement and hence show that the use of the latency predictors does not lead to any deterioration in overall performance.

\bibliography{iclr2021_conference}
\bibliographystyle{iclr2021_conference}

[1] Cai H, Gan C, Wang T, Zhang Z, Han S. Once-for-all: Train one network and specialize it for efficient deployment. arXiv preprint arXiv:1908.09791. 2019 Aug 26.

[2] Elsken T, Metzen JH, Hutter F. Neural Architecture Search. In Journal of Machine Learning Research, 2019.

[3] Cai H, Zhu L, Han S. Proxylessnas: Direct neural architecture search on target task and hardware. arXiv preprint arXiv:1812.00332. 2018 Dec 2.

[4] Wang Q, Chu X. GPGPU performance estimation with core and memory frequency scaling. IEEE Transactions on Parallel and Distributed Systems. 2020 Jun 24;31(12):2865-81.

[5] Parakh AK, Balakrishnan M, Paul K. Performance estimation of GPUs with cache. In 2012 IEEE 26th International Parallel and Distributed Processing Symposium Workshops \& PhD Forum 2012 May 21 (pp. 2384-2393). IEEE.

[6] Issa J, Figueira S. A performance estimation model for gpu-based systems. In 2012 2nd International Conference on Advances in Computational Tools for Engineering Applications (ACTEA) 2012 Dec 12 (pp. 279-283). IEEE.

[7] Liaw Richard, Liang Eric, Nishihara Robert, Moritz Philipp, Gonzalez E. Joseph, Stoica Ion. Tune: A Research Platform for Distributed Model Selection and Training. In ICML AutoML Workshop, 2018.

[8] Han Song, Pool Jeff, Tran John, Dally J. William. Learning both Weights and Connections for Efficient Neural Networks. In NIPS, 2015.



\clearpage
\appendix
\section{Latency Predictors for Once-For-All NAS}
\label{Appendix:nas}
\begin{figure}[h]
    \centering
    \subfigure[Samsung Galaxy Note10 using Fixed-Kernel CompOFA]{\includegraphics[width=13.5cm]{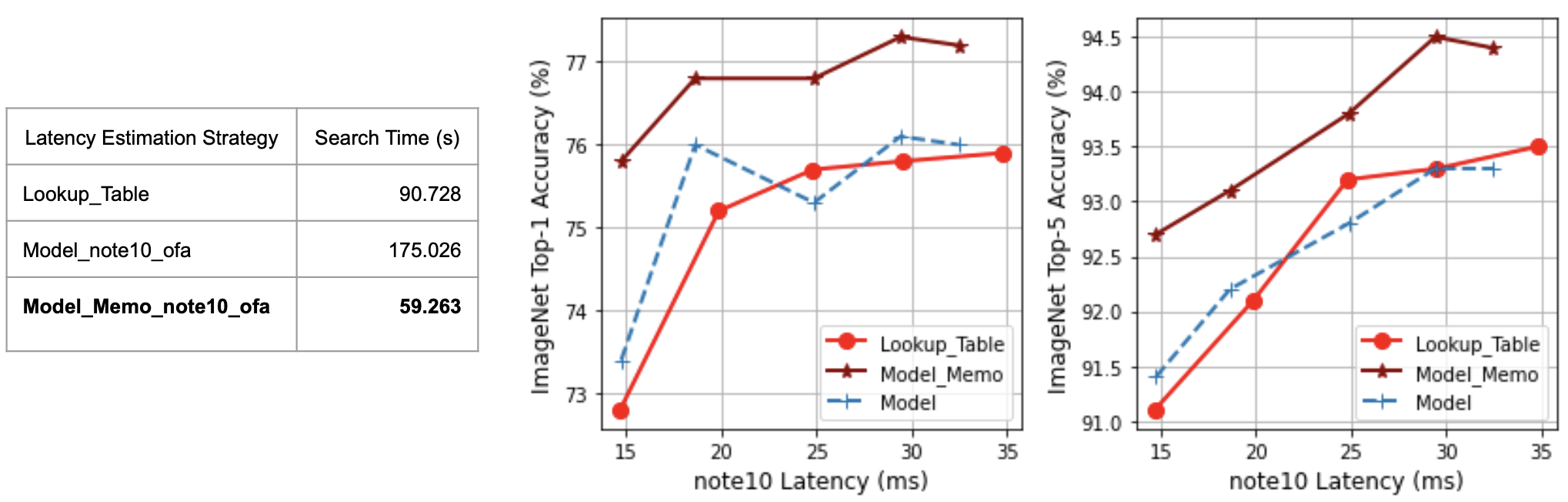}}
    \subfigure[Samsung Galaxy Note10 using Elastic-Kernel CompOFA]{\includegraphics[width=13.5cm]{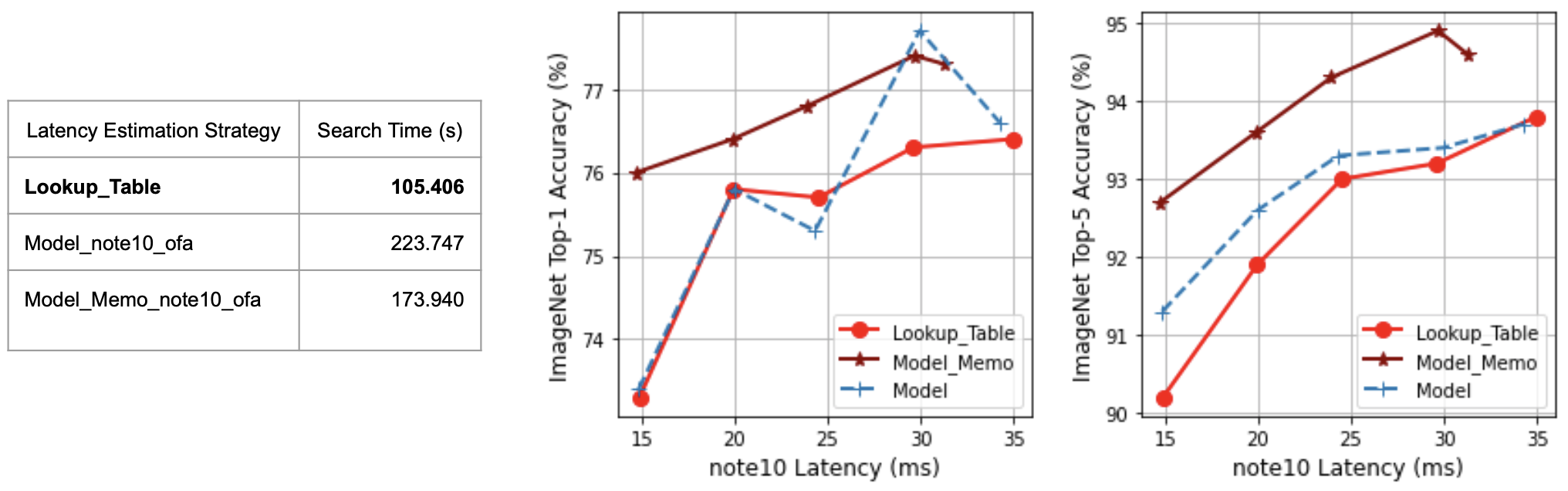}}
    \subfigure[RTX 2080 Ti GPU using Fixed-Kernel CompOFA]{\includegraphics[width=13.5cm]{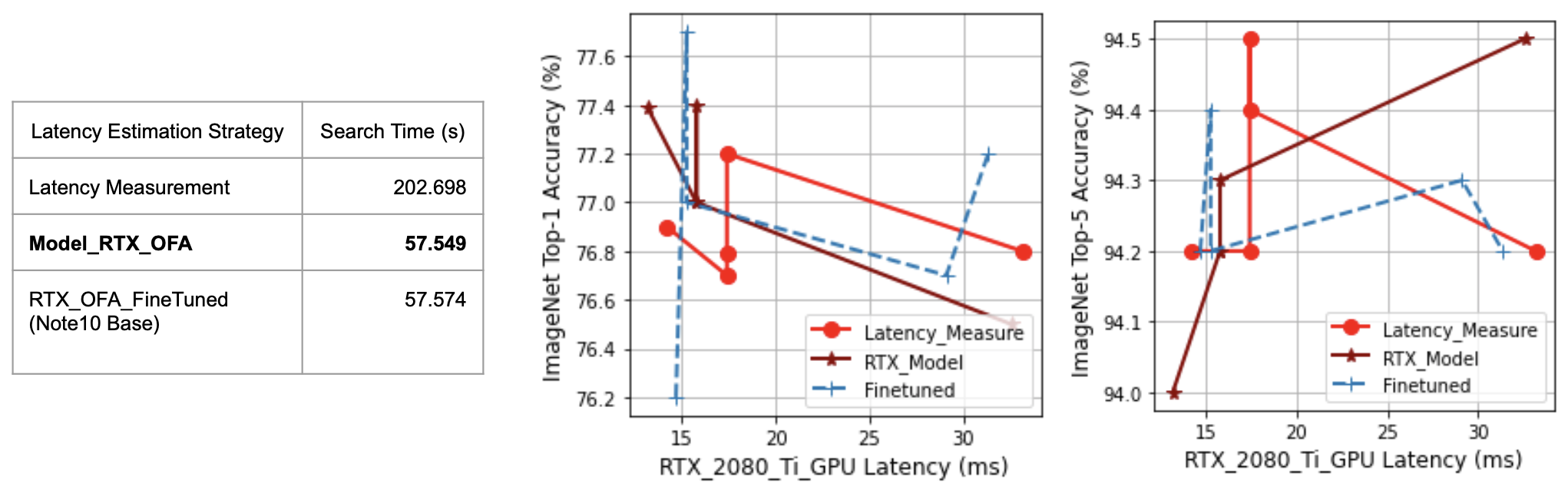}}
    \subfigure[RTX 2080 Ti GPU using Elastic-Kernel CompOFA]{\includegraphics[width=13.5cm]{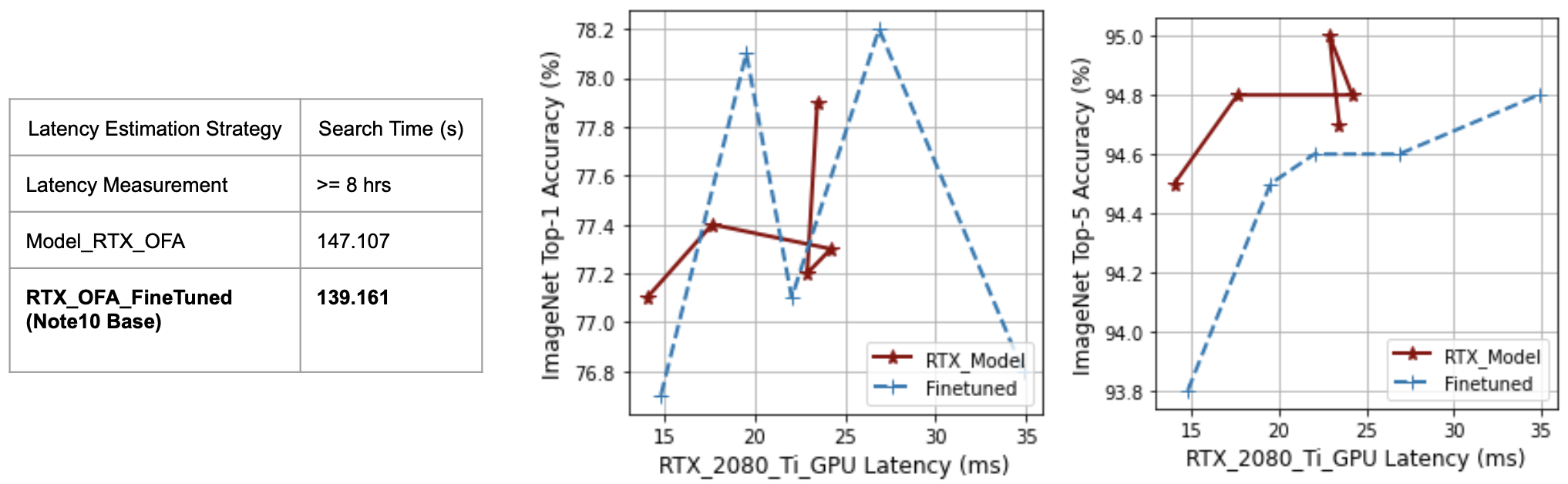}}

\caption{NAS experiment results for search time, accuracy, and latency}
\label{fig:nas_application}
\end{figure}

\end{document}

%% file: math_commands.tex
\usepackage{amsmath,amsfonts,bm}

\def\eqref#1{equation~\ref{#1}}

\def\1{\bm{1}}

\DeclareMathAlphabet{\mathsfit}{\encodingdefault}{\sfdefault}{m}{sl}
\SetMathAlphabet{\mathsfit}{bold}{\encodingdefault}{\sfdefault}{bx}{n}

